\documentclass[a4paper,twoside]{article}

\usepackage{epsfig}
\usepackage{subcaption}
\usepackage{calc}
\usepackage{amssymb}
\usepackage{amstext}
\usepackage{amsmath}
\usepackage{amsthm}
\usepackage{multicol}
\usepackage{pslatex}
\usepackage{apalike}
\usepackage{SCITEPRESS}     

\begin{document}

\title{Can we detect harmony in artistic compositions?\\A machine learning approach}

\author{\authorname{Adam Vandor\sup{1}, Marie van Vollenhoven\sup{2}, Gerhard Weiss \sup{1} and Gerasimos Spanakis\sup{1}\orcidAuthor{0000-0002-0799-0241}}
\affiliation{\sup{1}Department of Data Science and Knowledge Engineering, Maastricht University, Maastricht, Netherlands}
\affiliation{\sup{2}Infinity Games, www.infinitygames.xzy, Maastricht, Netherlands}

\email{vandoradam1@gmail.com, marie@vanhollenhoven.nl, \{gerhard.weiss,jerry.spanakis\}@maastrichtuniversity.nl}
}


\abstract{
Harmony in visual compositions is a concept that cannot be defined or easily expressed mathematically, even by humans. The goal of the research described in this paper was to find a numerical representation of artistic compositions with different levels of harmony. We ask humans to rate a collection of grayscale images based on the harmony they convey. To represent the images, a set of special features were designed and extracted. By doing so, it became possible to assign objective measures to subjectively judged compositions. Given the ratings and the extracted features, we utilized machine learning algorithms to evaluate the efficiency of such representations in a harmony classification problem. The best performing model (SVM) achieved 80\% accuracy in distinguishing between harmonic and disharmonic images, which reinforces the assumption that concept of harmony can be expressed in a mathematical way that can be assessed by humans.
}

\onecolumn \maketitle \normalsize \setcounter{footnote}{0} \vfill

\section{\uppercase{Introduction}}

Harmony is an abstract concept that does not have a formally precise definition. Depending on the domain, such as painting, music or architecture, harmony means something different to different persons. When we think of harmony, we associate it with a compilation of independent elements, which as a result create a consistent, pleasing arrangement. Even though there exist well-known patterns that carry harmonious sentiments in the eyes of the people, like the golden-ratio \cite{di2007golden} in images, the inherent source of their harmonic nature is not obvious. The research described here aimed at exploring whether the individual perception of harmony in images can be expressed in mathematical terms. Specifically, our research aims at addressing the question whether it is possible to use machine learning techniques to generate a mathematically founded model of a person's subjective understanding of harmony. 

Our hypothesis is that if machine learning models are able to be trained in order to confidently predict the labels one would assign to the compositions, then it would mean that there exists some numerical representation of a composition that reflects its harmonic level. Therefore, we need to extract features from the compositions that can further be used by machine learning algorithms. The main difficulty of this approach lies in the designing of such features that carry general and meaningful information about the compositions. We also need to take into consideration that no classical data augmentation techniques - such as rotation, translation, flipping, etc. - can be applied in the research, as the resulting images would be completely new data points that might be assessed differently.

After presenting a short overview of the state-of-the-art techniques in Section \ref{sect:relatedwork}, Section \ref{sect:methodology} outlines the dataset and the process of feature extraction used in our research, and describes the necessary transformations of the data and how the uncertain nature of the target variables was handled, followed by the machine learning framework used, and Section \ref{sect:results} shows the experimental results achieved through machine learning. 

\section{\uppercase{Related Work}}
\label{sect:relatedwork}

On the technical side, visual feature extraction and image classification are broadly investigated topics in the field of Artificial Intelligence and Computer Vision. Research and applications regarding object recognition cover several domains. In \cite{wu2007leaf} methods for leaf recognition are examined. In \cite{lecun1990handwritten} the classification of handwritten digits is explored using back-propagation. There are a number of other use cases, which consider computer vision based approaches in order to analyze and identify meaningful patterns. In \cite{zhao1998discriminant} techniques for face recognition are presented, in \cite{bharati2004image} the analysis of surface textures are investigated in order to assess the quality of produced goods, and in \cite{deepa2011survey} a collection of Artificial Intelligence based approached are surveyed for medical image classification. 

Feature extraction methods play indispensable roles in efficient image classification. Feature extraction algorithms such as SIFT \cite{lowe1999object} and SURF \cite{bay2006surf} introduce techniques to find generally meaningful patterns in images. \cite{nixon2012feature} covers a broad range of feature extraction and image processing techniques, and \cite{ping2013review} reviews feature extraction and representation techniques for further image classification.

On the philosophical side, the question that lies is whether an abstract concept like harmony can be expressed by quantitative measures. On the one hand, one argument, originating from Plato \cite{pappas2008plato} is that concepts like beauty can be associated with harmony, symmetry and unity. On the other hand, research suggests that there are also emotional processes that contribute to assessing aesthetics \cite{leder2004model}.

When it comes to evaluating artistic creations of computer programs, researchers have attempted to measure the aesthetic value using a ratio of the perceived order over complexity \cite{davis1936evaluation}. However, most of the work focuses on describing concepts like computational creativity and how it can be assessed \cite{jordanous2012standardised} but only a few approaches exist that try to quantify the concept of harmony in compositions, e.g. in \cite{salleh2015quantifying} authors are assessing the aesthetic quality of trochoids.

In this paper, a set of specific features is introduced, which is directly applicable to the problem of capturing the concept of harmony by carrying information about the arrangement of artistic compositions. Our expectation is that if a machine learning model can be trained to identify whether a composition is harmonic or not, then that would be a first step to build such a measure for quantifying such an abstract concept.

\begin{figure*}[h]
\centering
\includegraphics[width=\linewidth]{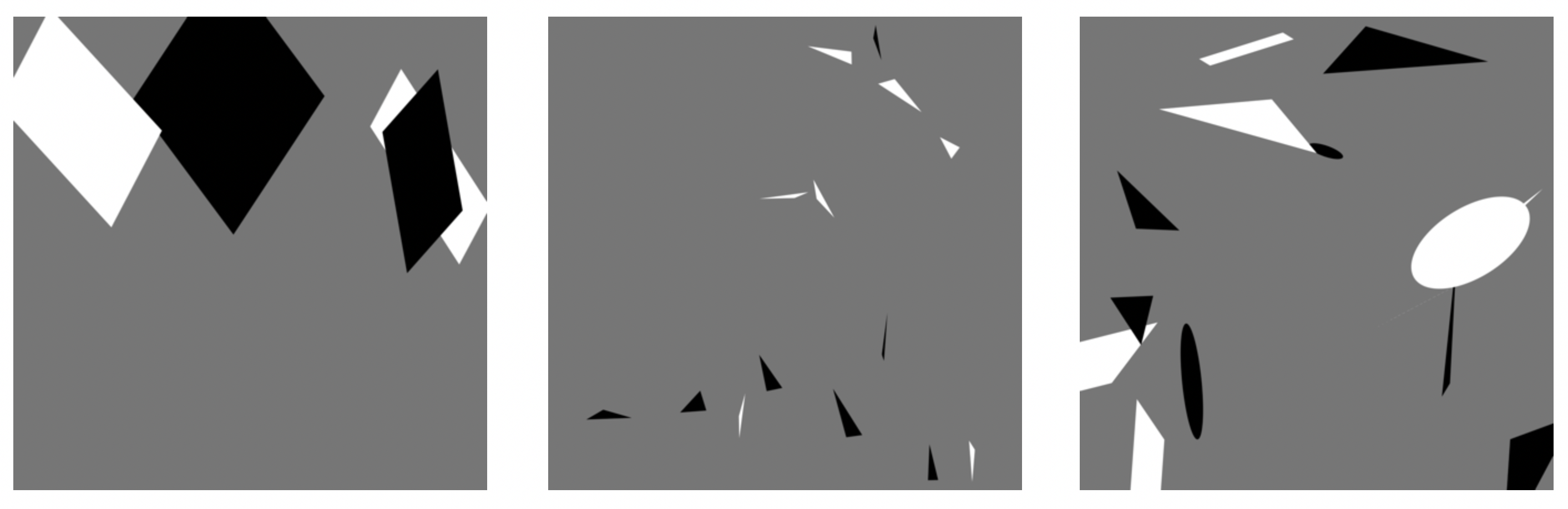}
\caption{Example compositions from the dataset}
\label{comps}
\end{figure*}

\section{\uppercase{Methodology}}
\label{sect:methodology}

\subsection{Dataset collection}
The dataset used in this research consists of a number of
randomly generated visual compositions by the application 'The Composition Game' \footnote{The Composition Game is a concept of artist Marie van Vollenhoven and is created together with interaction designer Felix Herbst.}. Each composition displays a set of black and white shapes on a square gray background. The position and rotation of each shape are randomly generated. The size of each shape is randomly drawn from a preset continuous range. The amount of black and white shapes, and the amount of circles, rectangles and triangles are set prior to the image generation. Figure \ref{comps} shows a few examples.

The participants' task was to evaluate each composition and to assign a number to it from a discrete range from 1 to 5, which they thought expresses the level of harmony of the image the most, where 5 means very harmonic and 1 means very disharmonic. We asked one participant to rate 8909 different compositions and we refer to this collection of compositions and ratings as \textit{the dataset}. In the future, we plan to include data from more participants as the study is ongoing.

\subsection{Feature Extraction}

In order to represent each composition as a vector in an $n$-dimensional space, we need to assign $n$ numerical values to it. The resulting feature vector is the concatenation of the extracted values and each component of the vector corresponds to a feature. 

One of the issues that arises is that some extracted features are highly dependent on the number of shapes, which could result in feature vectors having different lengths. For example, if a feature to be used is given by the distances between each shape in an image and the center of the image, then this feature would consist of three values in case there are three shapes in the image wheras it would consist of four values in case the image contains four shapes. 
For this reason, we use statistical properties (minimum, maximum, mean and standard deviation) of such features rather than the individual values themselves. In this way, the actual information related to such a feature is not represented in a straightforward form by enumerating all individual values, but through its statistical counterpart that encodes information in a compressed way. This representation still carries meaningful information and keeps the overall database manageable.

A summary of the designed features that had been extracted from the compositions is presented below.

\subsubsection{Number of shapes}

The feature returns the number of shapes in a composition.

\subsubsection{Number of specific shapes}

The feature returns the number of triangles, circles, rectangles and indeterminable shapes. Indeterminable shapes may appear on compositions when two or more shapes overlap.

\subsubsection{Number of specific colors}
The feature returns the number of black and the number of white shapes. 

\begin{figure}[h]
\centering
\scalebox{0.8}{
\includegraphics[width=1.0\linewidth]{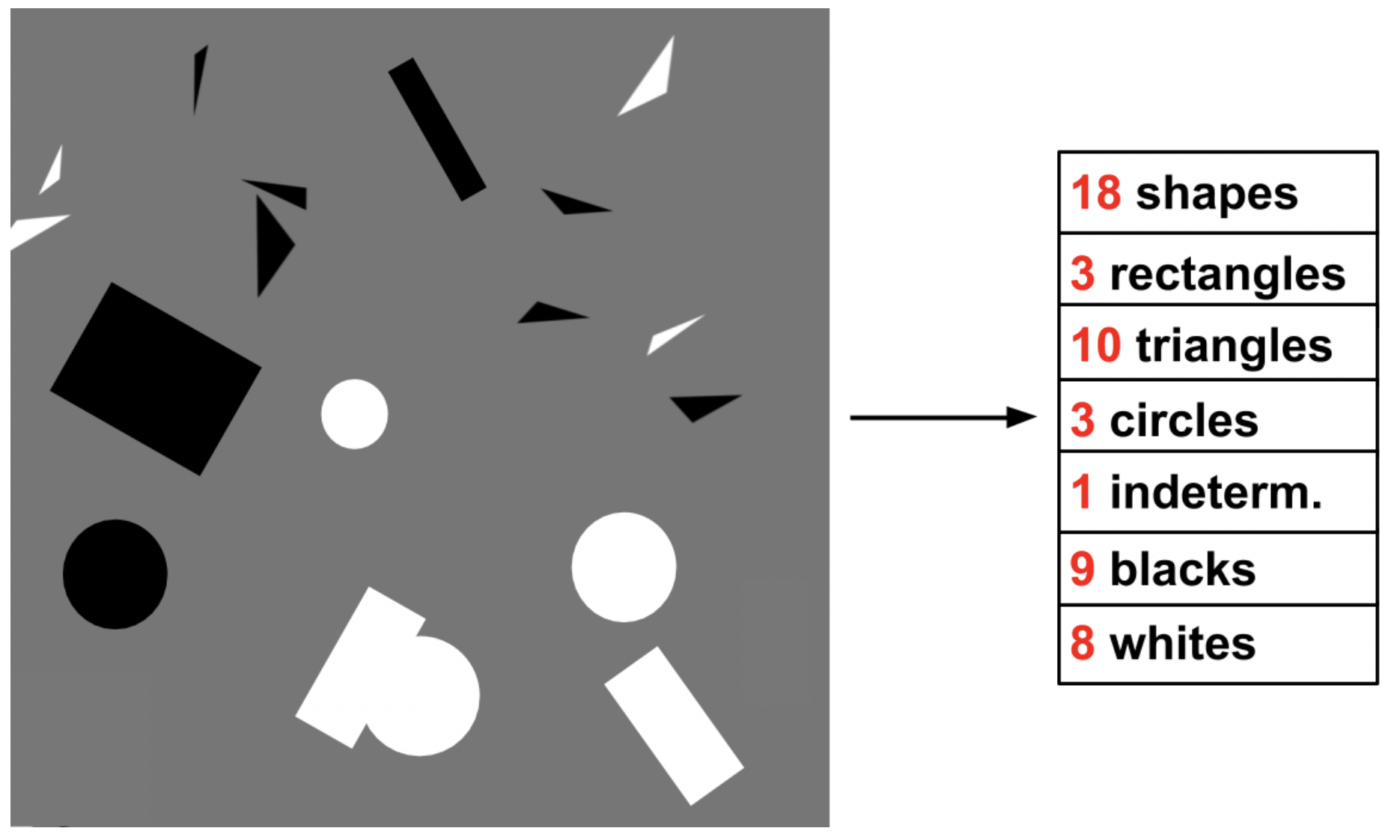}}
\caption{Features 1, 2 and 3}
\label{rawfeatures}
\end{figure}

\subsubsection{B\&W ratio}
The feature returns the ratio between the number of black and the number of white shapes. The denominator of the fraction is always the greater number out of the two. If one of the numbers is zero, the function returns zero.

\subsubsection{Number of groups}
\begin{figure}[h]
\centering
\scalebox{0.9}{
\includegraphics[width=1.0\linewidth]{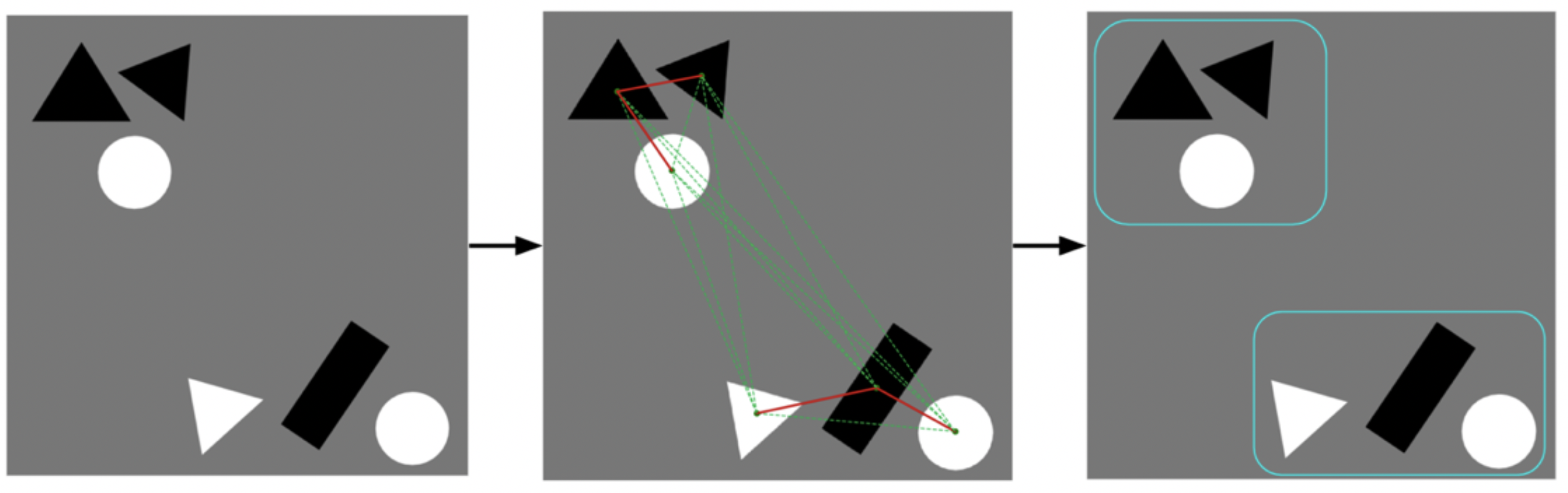}}
\caption{Grouping (Feature 5)}
\label{grouping}
\end{figure}

The feature returns the number of groups in a composition. Groups are subsets of all shapes in which the shapes are the closest to each other. The function first determines the centers of the shapes, then calculates the Euclidean distance between every possible pair. After this step, let’s consider the image as a graph. Each shape (vertex) is then connected to its closest neighbor. The number of groups in the composition is the number of disconnected  sub-graphs in the graph (see Figure \ref{grouping}).

\subsubsection{Covered area}
The features returns the ratio between the area covered by the shapes and the area of the image.

\subsubsection{Area covered by groups}
The feature determines the ratio between the area covered by each group and the area of the image. Then the statistical properties of the obtained values are extracted as mentioned above. 

\subsubsection{Entropy}
\begin{figure}[ht]
\centering
\scalebox{0.9}{
\includegraphics[width=1\linewidth]{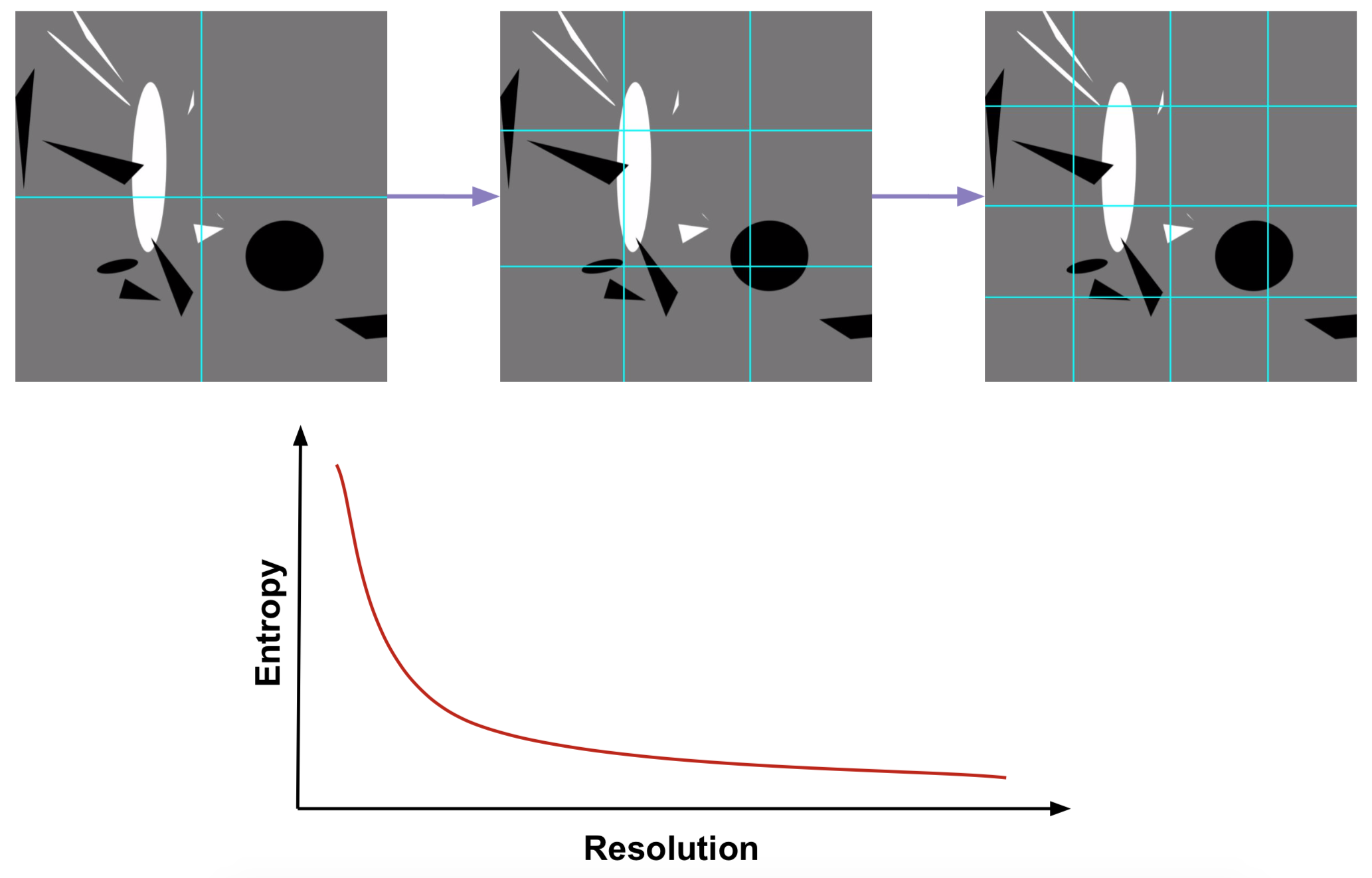}}
\caption{Entropy (Feature 8)}
\label{entropy}
\end{figure}

The feature returns three numbers that indicate how much the shapes are spread in a composition; the more spread they are, the higher its entropy (see Figure \ref{entropy}). By decreasing the sizes of the squares in each step, the function fits  square grids on the image multiple times. After every iteration, the number of gird cells that contain non-gray pixels are determined. This number is then divided by number of grid cells in the current iteration. The result of these divisions are saved after each step. If we then plot the values, we get a curve which shows how the entropy decreases over the iterations. In the last step, a second degree polynomial is fit to the curve. The function returns the values of \textit{a, b} and \textit{c} determined for the polynomial
\begin{equation}
ax^2+bx+c \end{equation}

\subsubsection{Bounding}
\begin{figure}[htb]
\centering
\scalebox{0.8}{
\includegraphics[width=1.0\linewidth]{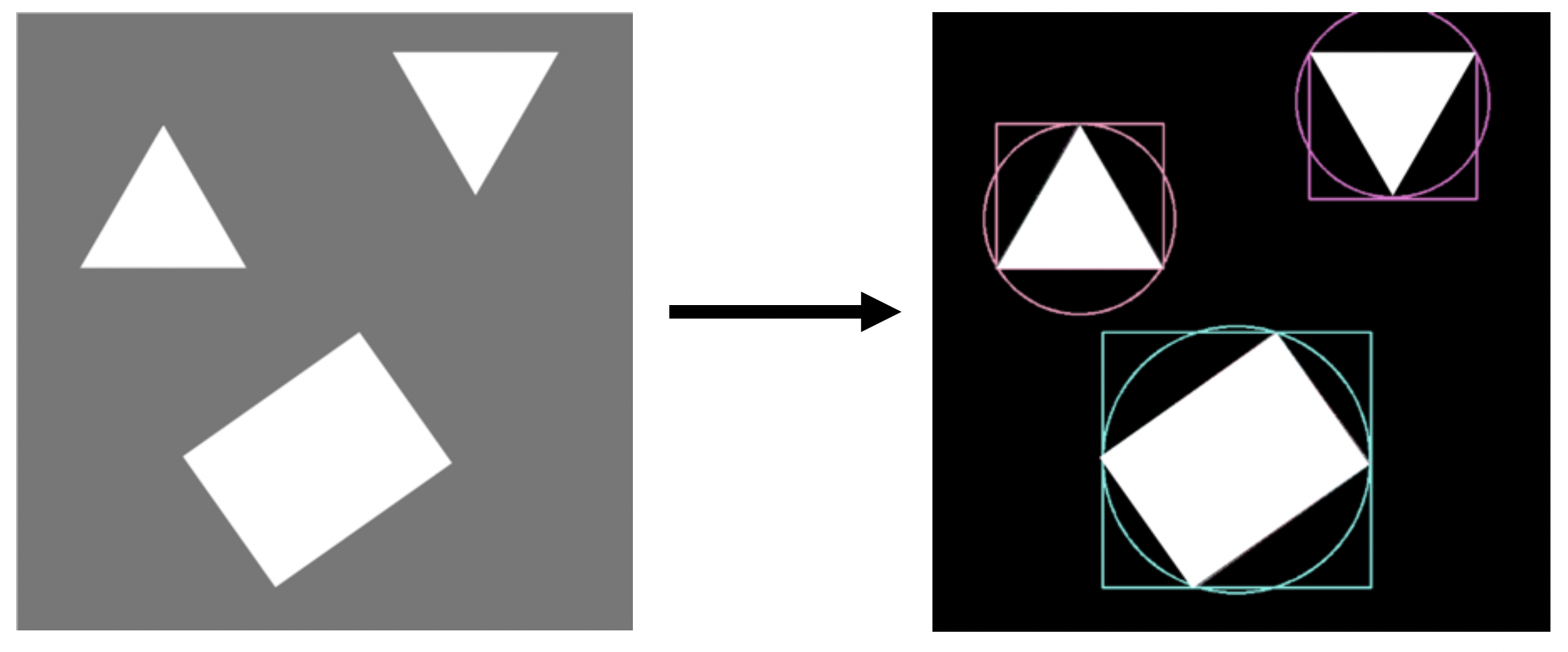}}
\caption{Bounding (Feature 9)}
\label{bounding}
\end{figure}

The feature first fits a bounding circle and a bounding rectangle around every shape in a composition (see Figure \ref{bounding}. The radiuses of the bounding circles, and the widths and heights of the bounding rectangles are stored. The function determines the statistical properties of the list of radiuses, widths, heights, width/heights and width*heights.

\subsubsection{Color distribution}
\begin{figure}[htb]
\centering
\centering
\scalebox{0.8}{
\includegraphics[width=1.0\linewidth]{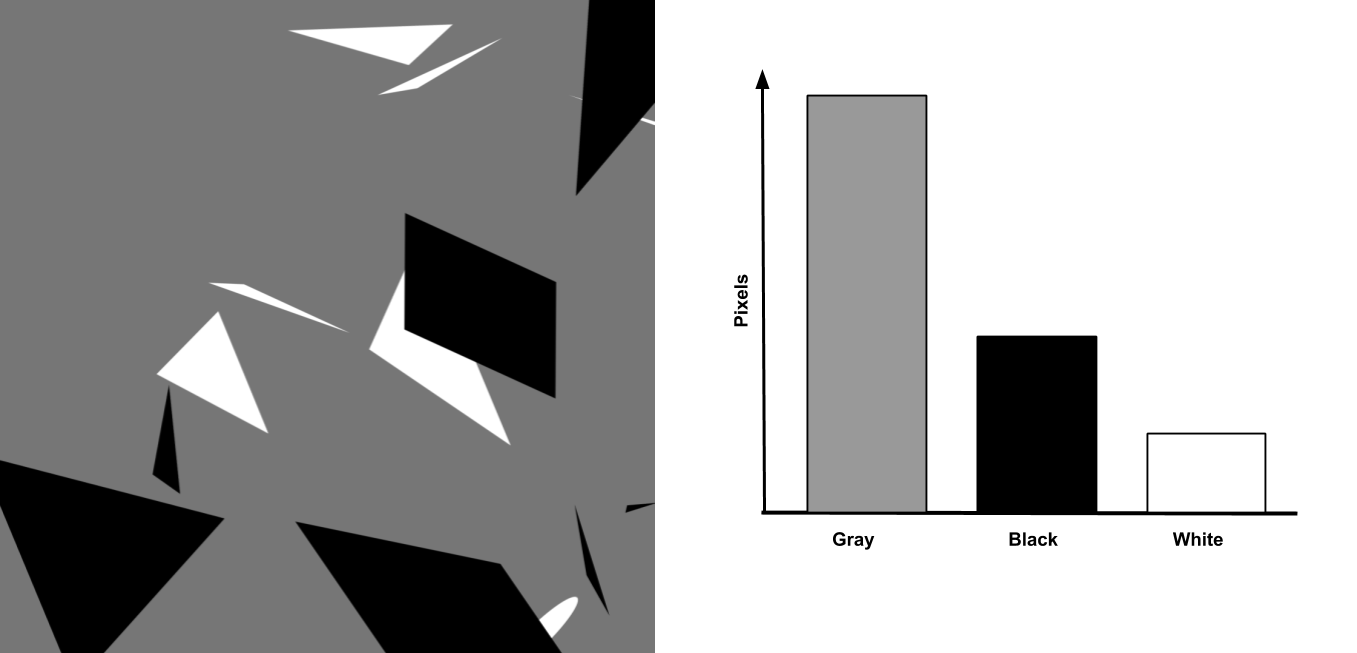}}
\caption{Color distribution (Feature 10)}
\label{colordist}
\end{figure}

This feature returns the number of gray, black and white pixels in an image (see Figure \ref{colordist}).

\subsubsection{Two-third points}

\begin{figure}[htb]
\centering
\scalebox{0.8}{
\includegraphics[width=1.0\linewidth]{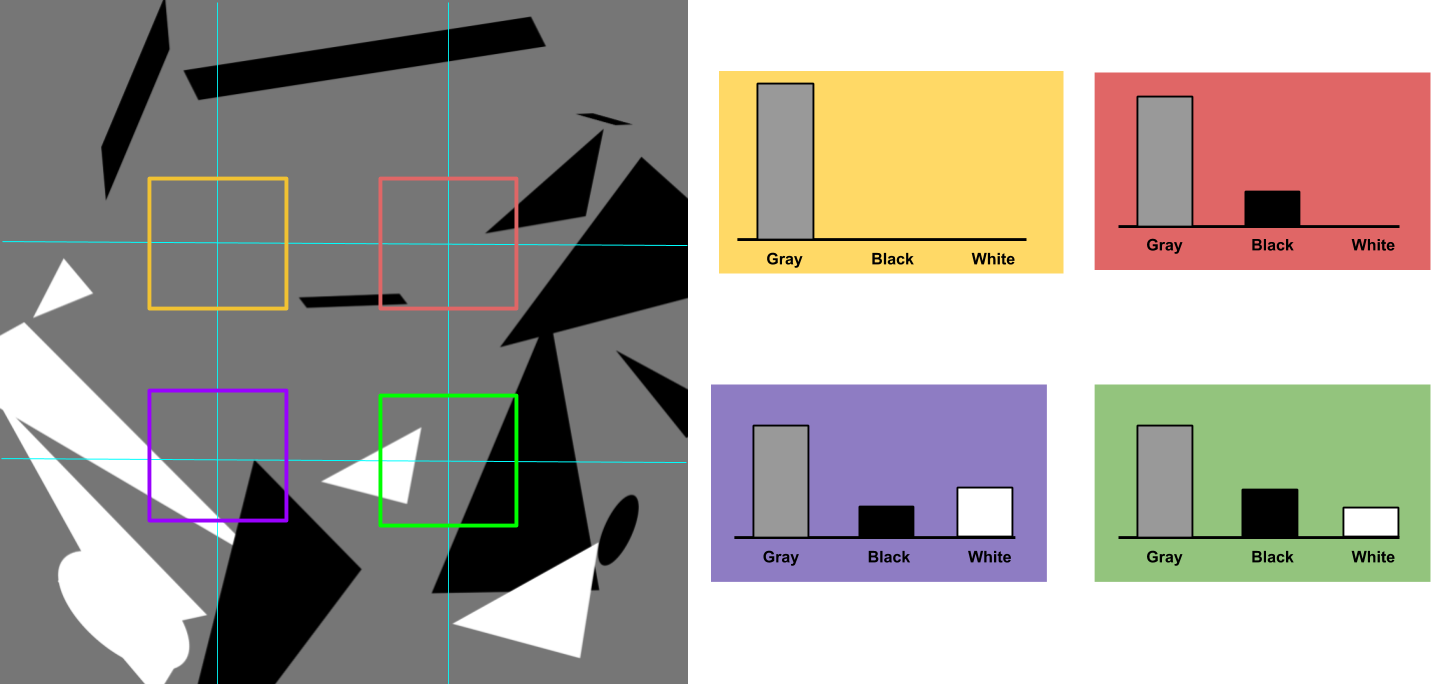}}
\caption{Two-third points (Feature 11)}
\label{twothird}
\end{figure}

This feature divides the image plane into thirds along both the horizontal and the vertical axes and analyzes the surrounding of those 4 points where the aforementioned lines intersect (see Figure \ref{twothird}). The surrounding is defined as a square whose center is an intersection point itself. The function returns the color distribution of these four areas. The motivation behind the feature is that the attention of the spectator is mostly drawn to the surrounding of these points \cite{amirshahi2014evaluating}.

\subsubsection{Balance}
\begin{figure}[htb]
\centering
\scalebox{0.8}{
\includegraphics[width=1.0\linewidth]{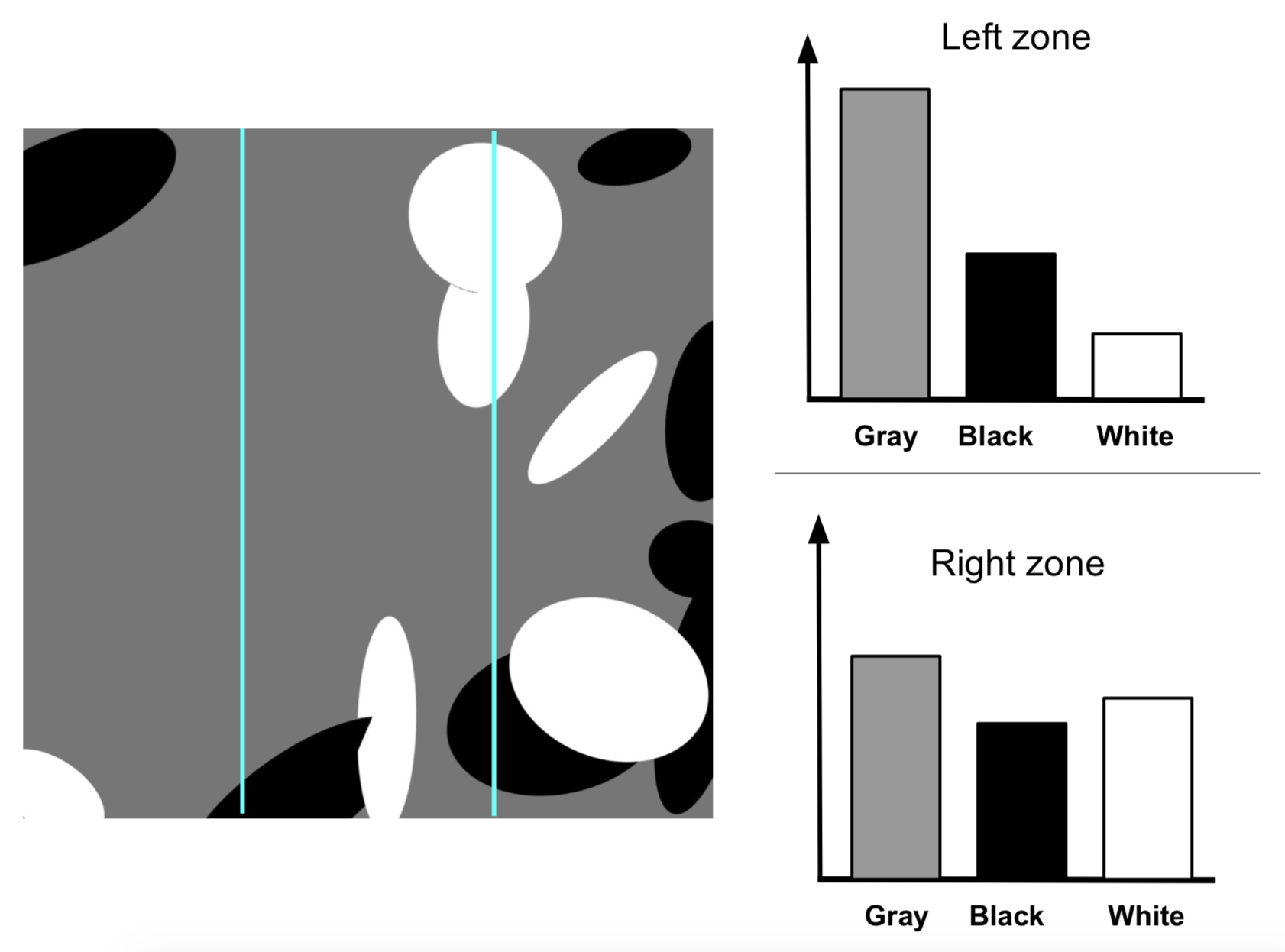}}
\caption{Balance (Feature 12)}
\label{balance}
\end{figure}

This feature determines the color distribution in the left and the right third of the image. This is an indicator about how well the two sides of the composition are balanced (see Figure \ref{balance}).

\subsubsection{Gravity}
\begin{figure}[htb]
\centering
\scalebox{0.5}{
\includegraphics[width=1.0\linewidth]{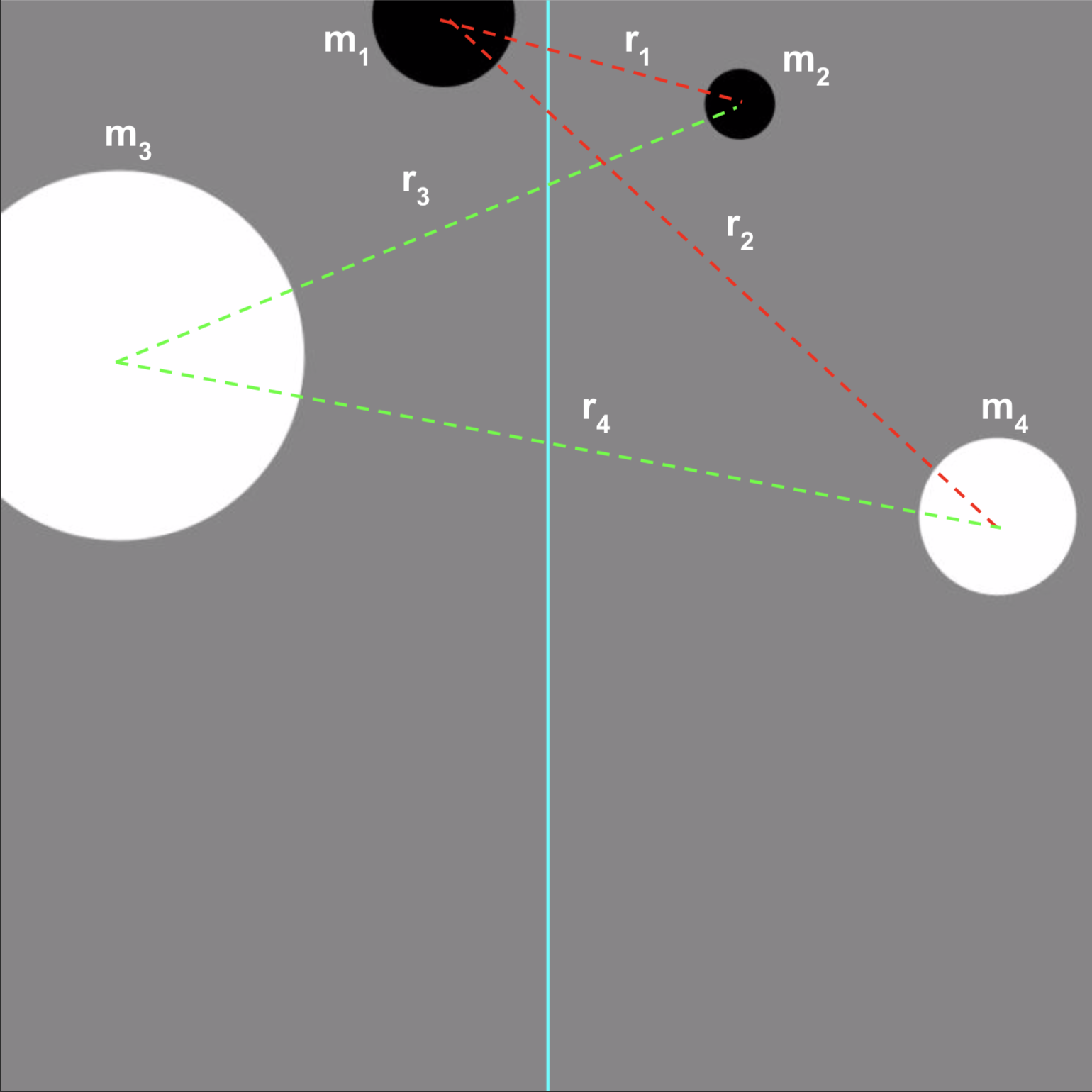}}
\caption{Gravity (Feature 13)}
\label{gravity}
\end{figure}

This feature calculates how much the pairs of shapes on left and the right side of the composition ``pull" each other, which is another type of indication of balance in the overall picture. First, the composition is split into two equal halves, then the center and area of the each shape are determined. According to Newton’s law \cite{newton1987philosophiae}, the occurring gravitational force between two arbitrary shapes –- one belonging to the right plane and the other to the left one -- is computed. The gravitational equation is expressed as

\begin{equation}
    F = \gamma \cdot \frac{m_{1} \cdot m_{2}}{r}
\end{equation}
\label{newton}

\noindent where $F$ denotes the force between objects $m_{1}$ and $m_{2}$, given their spatial distance $r$. $\gamma$ is the gravitational constant, which in this paper is set to $10^{-8}$.
The \textit{mass} of a shape is interpreted as its area (see Figure \ref{gravity}).

\subsubsection{Areas}
This feature return the statistical properties of the list of areas of the shapes in a composition.

\subsubsection{Shape-center distance}
\begin{figure}[htb]
\centering
\scalebox{0.5}{
\includegraphics[width=1.0\linewidth]{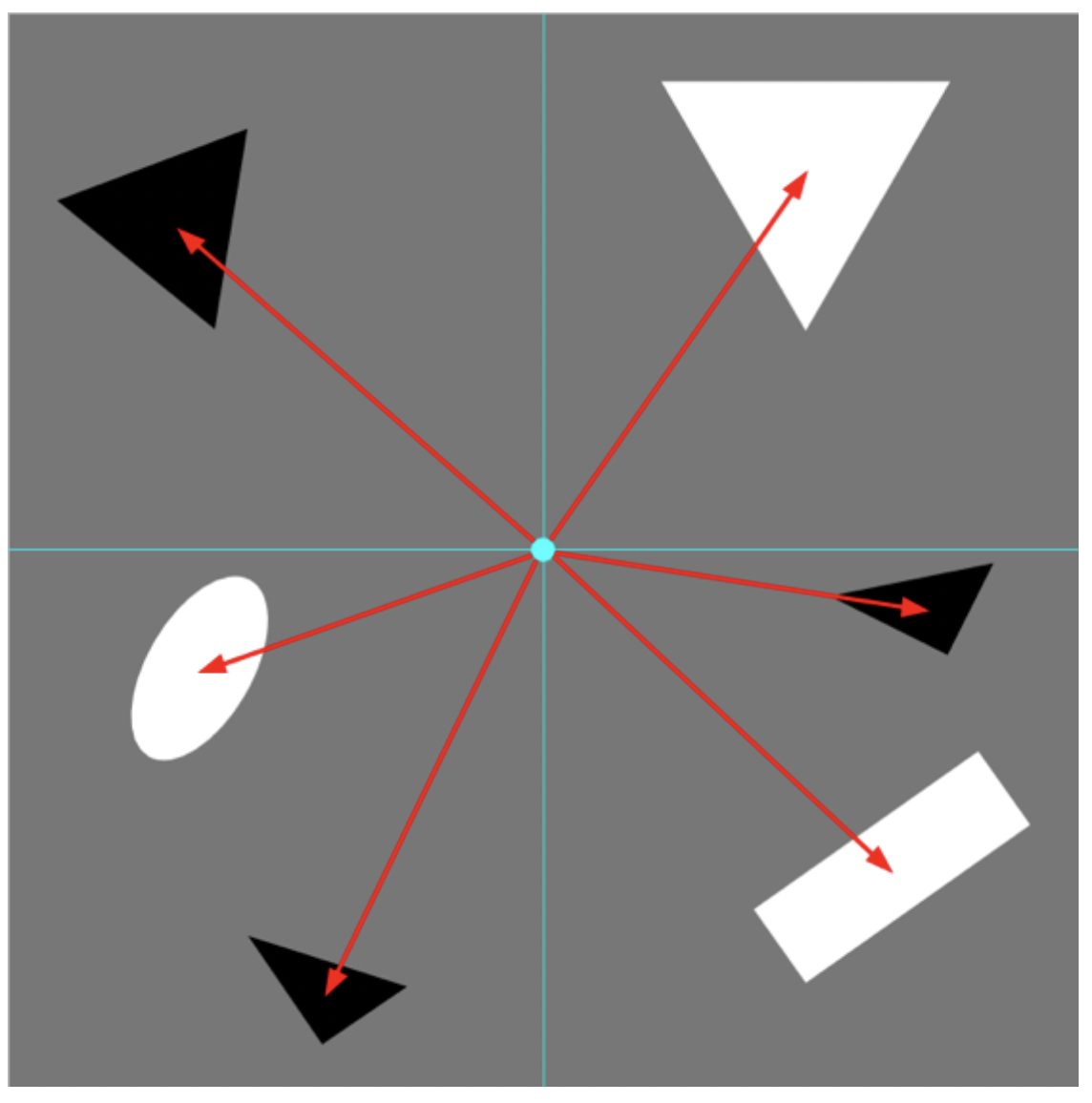}}
\caption{Shape-center distance (Feature 15)}
\label{ceneterdist}
\end{figure}

This feature calculates the distances between the center of each shape and the center of the image, then returns the statistical properties of these values (see Figure \ref{ceneterdist}).

\subsubsection{SURF features}
In order to obtain a more general representation of a composition, traditional computer vision techniques had been applied. After extracting the SURF features \cite{bay2006surf} from the images, $k$ = 5, 10, 20, 50, 100, 200 and 500 clusters had been created with k-means. The main idea of SURF is to extract detectors and descriptors from images which are not susceptible to rotation, scaling etc. Using the visual bag of words approach, each row of the extracted SURF matrices had been assigned to its nearest cluster center. This way, each composition receives a unique, histogram-like representation with $k$ bars. 

\subsubsection{Convolutional Autoencoder}
\begin{figure*}[htb]
\centering
\scalebox{0.85}{
\includegraphics[width=1\linewidth]{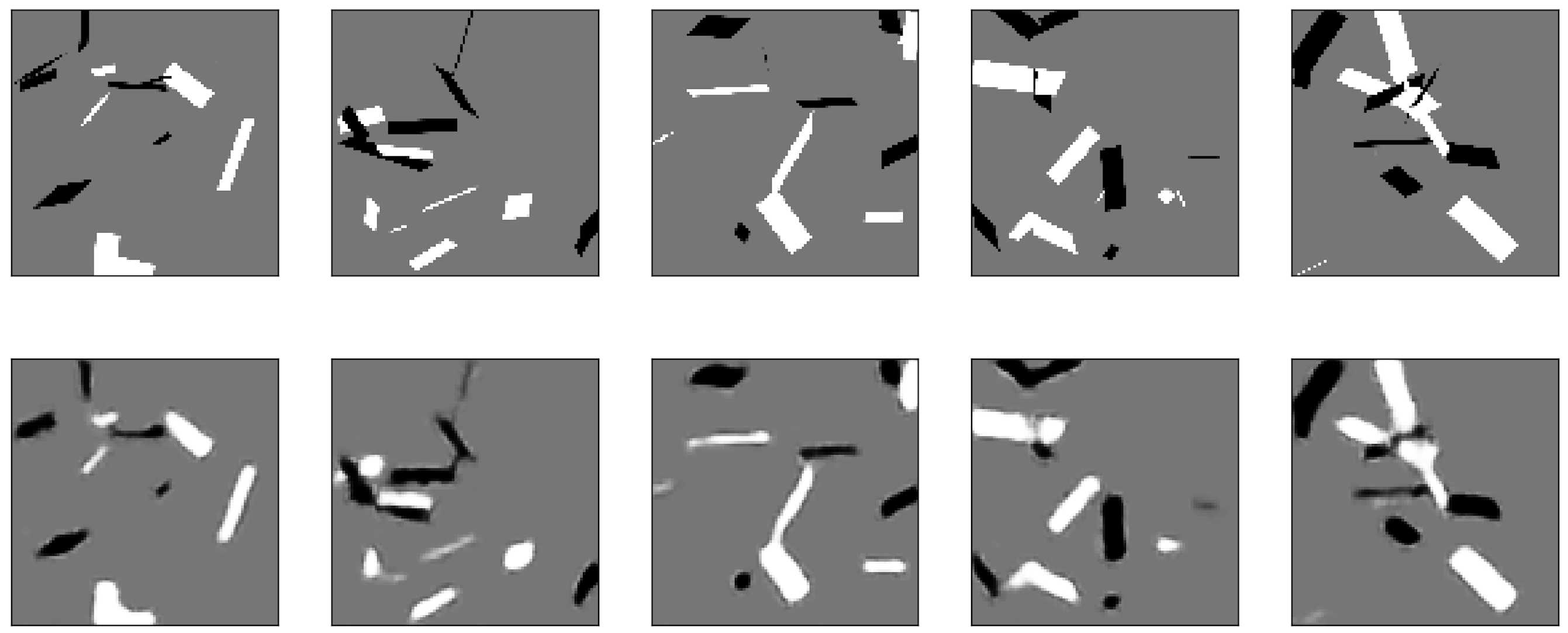}}
\caption{Original (top) and reconstructed (bottom) images}
\label{CAE}
\end{figure*}

To obtain a more dense representation of the compositions, a Convolutional Autoencoder \cite{masci2011stacked} had been trained on 70\% of the dataset. If the reconstructed images look like the original ones, it means that the encoding layer does carry sufficient information about the compositions, thus the compressed form of an image can be used as a new feature. For this purpose, the images had been resized to 100 x 100 pixels. Figure \ref{CAE} shows some reconstructed images from the test set. 
The results indicate that the encoding layer does carry enough information about the original images and can be used to generate new features for the training phase. The encoded images have the size of 13 x 13 pixels, which gives 169 new features for a composition.

\subsection{Pre-processing}
\subsubsection{Feature transformation}
In order to improve the performance of the models to be learnt, the distribution of every feature in the dataset was checked and subjected to transformations if needed. Specifically, in case the distribution of a feature over the dataset looked like a skewed Gaussian, Box-Cox transform was applied; when a feature was dominated by a single value, square-rooting was used, and if the feature followed an exponential distribution, log transform was applied. In cases where a feature contained outliers, they were removed. After transforming the features, the dataset was normalized. 

\subsubsection{Extending the dataset}
In order to further smooth the dataset some additional features could be added by performing transformations on them. Given the polished dataset the following transformations were applied to it: Principal Component Analysis (PCA) \cite{jolliffe2011principal} with $n$ = 30 components, and truncated Singular Value Decomposition (SVD) \cite{golub1971singular} with $n$ = 9 components. The dataset was then extended by its own dense representations. Overall, each composition is represented as a 321 dimensional feature vector.

\subsubsection{Fixing the target classes}
\begin{figure}[htb]
\centering
\scalebox{1.0}{
\includegraphics[width=1\linewidth]{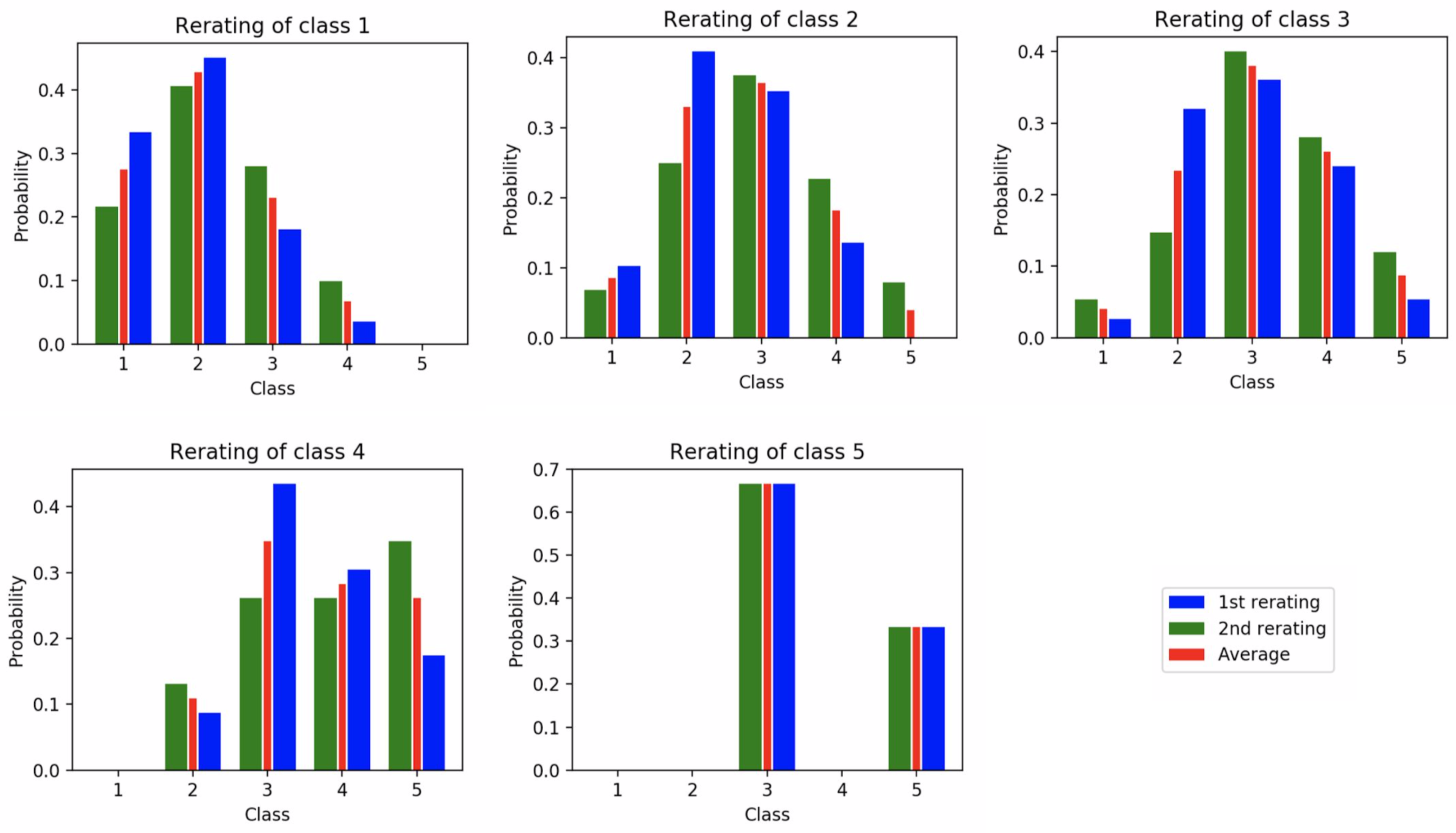}}
\caption{Re-ratings of participant}
\label{rerate_distr2}
\end{figure}

Given the highly subjective nature of the task, we cannot be certain that participants rate the harmonic nature of the compositions in a consistent way. This means that we cannot treat the ratings of the compositions as if they were perfectly expressing their harmonicity as they experience it. Determining the harmonicity of a composition may depend on many factors such the current mood of the participants, which means they might rate the same images differently in different situations. In order to smooth out this distortion, participant was asked to re-rate a subset of compositions two more times. Provided the re-ratings we can determine a more robust overall rating for each composition. Participant was asked to re-rate 300 compositions. Figure \ref{rerate_distr2} shows the deviation from each class after the re-ratings for the participant.

\begin{figure}[htb]
\centering
\includegraphics[width=1\linewidth]{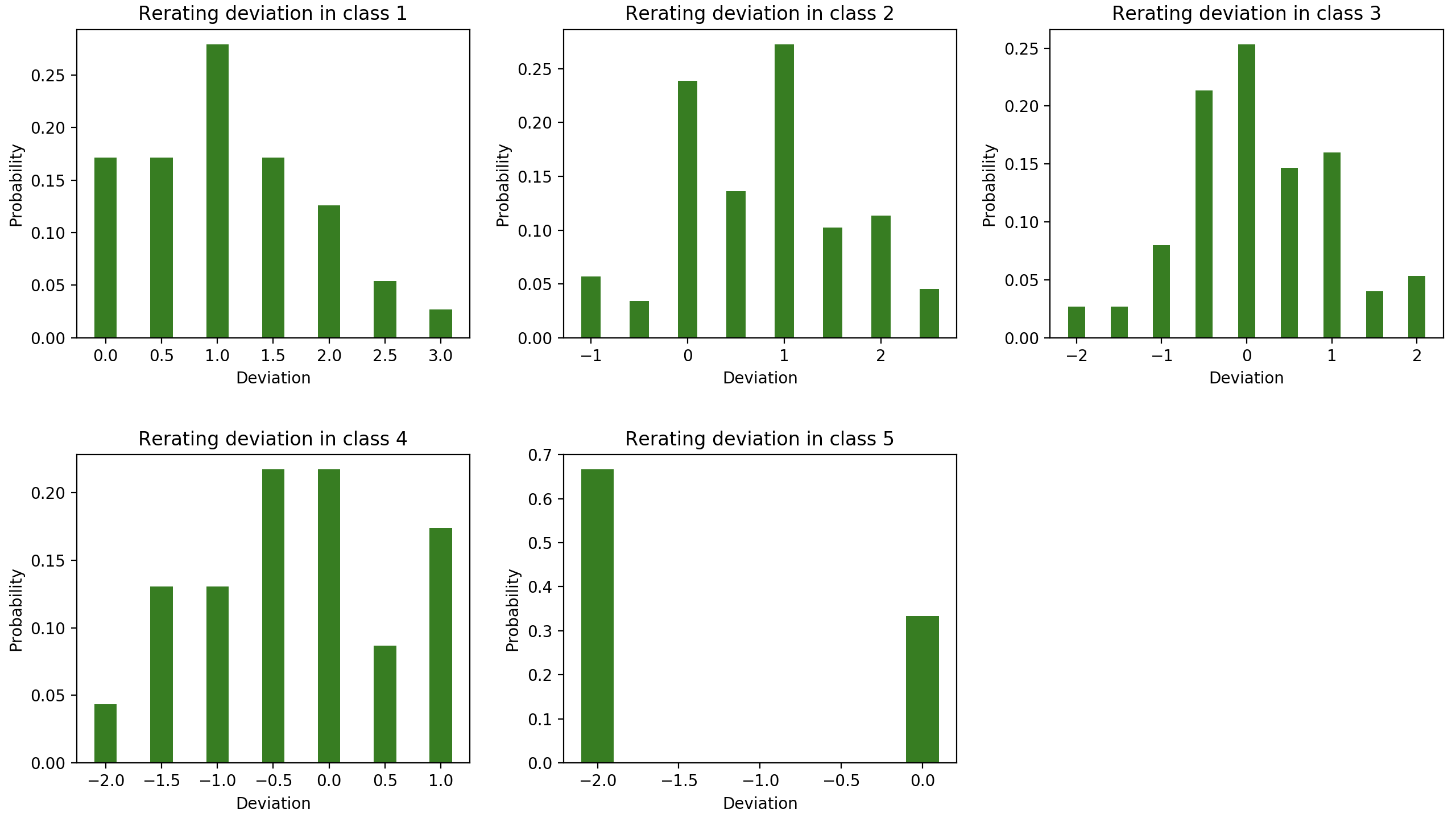}
\caption{Distributions of deviation of participant}
\label{fig:rerate_dev2}
\end{figure}

Given the re-ratings, it is possible to determine how much and how frequently participants deviate from their initial ratings. Figure \ref{fig:rerate_dev2} shows the distributions of deviation in each class. For example, in the second distribution in Figure \ref{fig:rerate_dev2}, we can see that the participant -- for the images initially rated as 2’s -- subtracts 1 unit around 5 percent of the time, subtracts 0.5 units around 3 percent of the time, preserves its rating around 25 percent of the time, adds 0.5 units around 15 percent of time, and so on.

By drawing samples from these distributions according to the appearing probabilities we can derive what average targets the compositions would get after several rounds of re-ratings \cite{bolthausen}. Table \ref{converged_values} shows the resulting values.

\begin{figure}[htb]
 \centering
 \scalebox{0.8}{
 \includegraphics[width=1\linewidth]{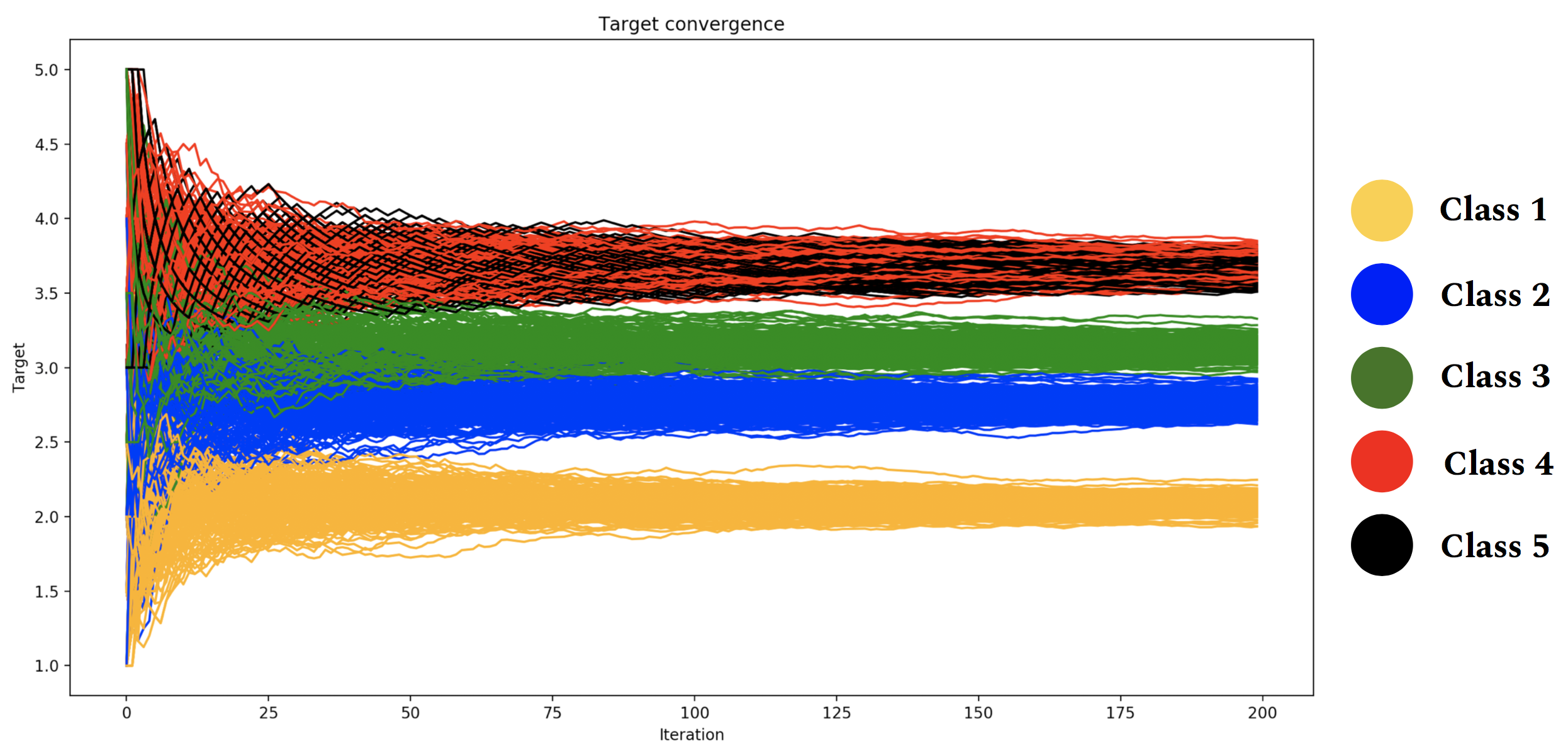}}
 \caption{Simulated convergence of ratings}
 \label{convergence_Jos}
 \end{figure}

The results of the simulation enable us to assign new values to the ratings, which express the harmonicity of compositions more reliably. Table \ref{converged_values} contains the old and the new ratings.

\begin{table}[htb]
\centering
\begin{tabular}{|c|c|}
\hline
\textbf{Old rating} & \textbf{New rating}\\
\hline
1 & 2.09\\
2 & 2.76\\
3 & 3.12\\
4 & 3.69\\
5 & 3.66\\
\hline
\end{tabular}
\caption{Converged values}
\label{converged_values}
\end{table}

From the converged values we saw that the updated ratings in classes [2, 3] and [4, 5] are fairly close to each other, so these two class pairs were merged to describe new categories. Thus, the original five classes are merged into three classes, expressing the level of harmony in a more compact way: $Bad$, $Neutral$ and $Good$. Table \ref{updated_classes} shows the final labels used for training.

\begin{table}[htb]
\centering
\begin{tabular}{|c|c|}
\hline
\textbf{Old class} & \textbf{New class} \\
\hline
1 & Bad\\
2 & Neutral\\
3 & Neutral\\
4 & Good\\
5 & Good\\
\hline
\end{tabular}
\caption{Mapping between old and new classes}
\label{updated_classes}
\end{table}

\subsection{Classification Approaches}
\label{sect:approach}

Given the final dataset with all 321 features per composition and a clearer notion about what the targets (classes) represent, we can proceed with applying a classification model on the dataset. We utilized state-of-the-art machine learning models stemming from different categories:

\begin{itemize}
    \item Random Forests \cite{breiman2001random} follow the bagging principle and construct multiple (shallow) decision trees during training time and they predict the class (in our case the rating) as the mode of predictions of all trees.
    
    \item Gradient Boosting \cite{friedman2001greedy} combines several weak classifiers (in our case decision trees). The whole idea of boosting is to build the final model step by step: each iteration of the model is based on a modified version of the original dataset with the goal to reduce the classification error of specific data points. We also employ XGBoost \cite{chen2016xgboost} is a fast and efficient implementation of gradient boosted decision trees.
    
    \item Logistic Regression \cite{bishop2006pattern} is one of the most widely used  classification techniques. We also employ Ridge classification \cite{hoerl1970ridge} which can address the multicollinearity issue that large feature spaces suffer from.
    
    \item Support Vector Machines \cite{cortes1995support} map the input space into separate categories divided by hyperplanes as wide as possible. Different kernels can be used in order to transform the non-linear input space into a higher dimensional linear one.  
    
    \item Multi-layer Perceptrons \cite{hinton1990connectionist} are also used as the simplest feed-forward artificial neural network (ANN) for classification.
\end{itemize}

In order to further improve the performance of the predictions, we combined the above models using a simple ensemble way (with a naive voting) \cite{dietterich2000ensemble} and using stacked generalization (which learns a new model to learn how to best combine the predictions) \cite{wolpert1992stacked}.

In the training phase, all four possible setups were explored with regards to the targets:

\begin{itemize}
\item \textbf{BN}: $Bad$ vs. $Neutral$,
\item \textbf{BG}: $Bad$ vs. $Good$,
\item \textbf{NG}: $Neutral$ vs. $Good$,
\item \textbf{BNG}: using $Bad$, $Neutral$ and $Good$ classes.
\end{itemize}

When training the models, each setup was tested with three different arrangements of the dataset:
\begin{itemize}
    \item \textbf{D1}: All features are included,
    \item \textbf{D2}: SURF features are omitted,
    \item \textbf{D3}: SURF and Convolutional Autoencoder features are omitted.
\end{itemize}

In all setups, 70\% of the dataset was used for training and the remaining 30\% for testing. We used 10-fold cross-validation for all experiments and we further used a validation set on the training set to tune the hyperparameters of all algorithms. Code and data will be made available upon paper acceptance.

\section{\uppercase{Experimental Results}}
\label{sect:results}

Table \ref{tab:results} summarizes the cross-validated results (accuracy and variance) of the best performing (for the sake of space) models in each setup. The best performance in each setup is highlighted.

\begin{table*}[htbp]
  \centering
  \caption{Experimental results for all training setups}
    \begin{tabular}{|c|c|c|c|c|c|c|c|}
    \hline
    \multicolumn{2}{|c|}{\textbf{ Training Setup }} & \textbf{ Single model } &  Var  & \textbf{ Ensemble } &  Var  & \textbf{ Stacking } &  Var   \\
    \hline
    & \textbf{    D1 } &  0.66 (XGBoost)  & 0.015 & 0.65  & 0.006 &  0.65 (Ridge)  & 0.022 \\
\textbf{BN}          & \textbf{    D2 } &  0.66 (GB)  & 0.014 & 0.66  & 0.007 & \textbf{ 0.67 (SVM) } & 0.035 \\
          & \textbf{    D3 } &  0.66 (XGBoost)  & 0.008 & 0.65  & 0.006 &  0.64 (XGBoost)  & 0.017 \\
    \hline

     & \textbf{    D1 } &  0.74 (XGBoost)  & 0.034 & 0.72  & 0.017 &  0.73 (Ridge)  & 0.06 \\
\textbf{BG}           & \textbf{    D2 } &  0.73 (XGBoost)  & 0.029 & 0.73  & 0.016 & \textbf{ 0.80 (SVM) } & 0.042 \\
      & \textbf{    D3 } &  0.71 (Ridge)  & 0.03  & 0.7   & 0.02  &  0.71 (Ridge)  & 0.058 \\
    \hline

   & \textbf{    D1 } &  0.59 (LR)  & 0.041 & 0.58  & 0.03  &  0.58 (SVM)  & 0.049 \\
\textbf{NG}        & \textbf{    D2 } &  0.57 (LR)  & 0.048 & 0.56  & 0.023 &  0.57 (SVM)  & 0.063 \\
         & \textbf{    D3 } &  0.57 (XGBoost)  & 0.019 & 0.58  & 0.022 & \textbf{ 0.60 (SVM) } & 0.073 \\
    \hline

   & \textbf{D1 } &  0.48 (XGBoost)  & 0.023 & 0.46  & 0.018 &  0.47 (Ridge)  & 0.066 \\
\textbf{BGN}     & \textbf{D2 } &  0.47 (GB)  & 0.029 & 0.48  & 0.011 & \textbf{ 0.50 (SVM) } & 0.038 \\
         & \textbf{D3 } &  0.48 (GB)  & 0.043 & 0.47  & 0.016 &  0.48 (SVM)  & 0.041 \\
    \hline

    \end{tabular}%
  \label{tab:results}%
\end{table*}%

\begin{figure}[htb]
\centering\includegraphics[width=0.8\linewidth]{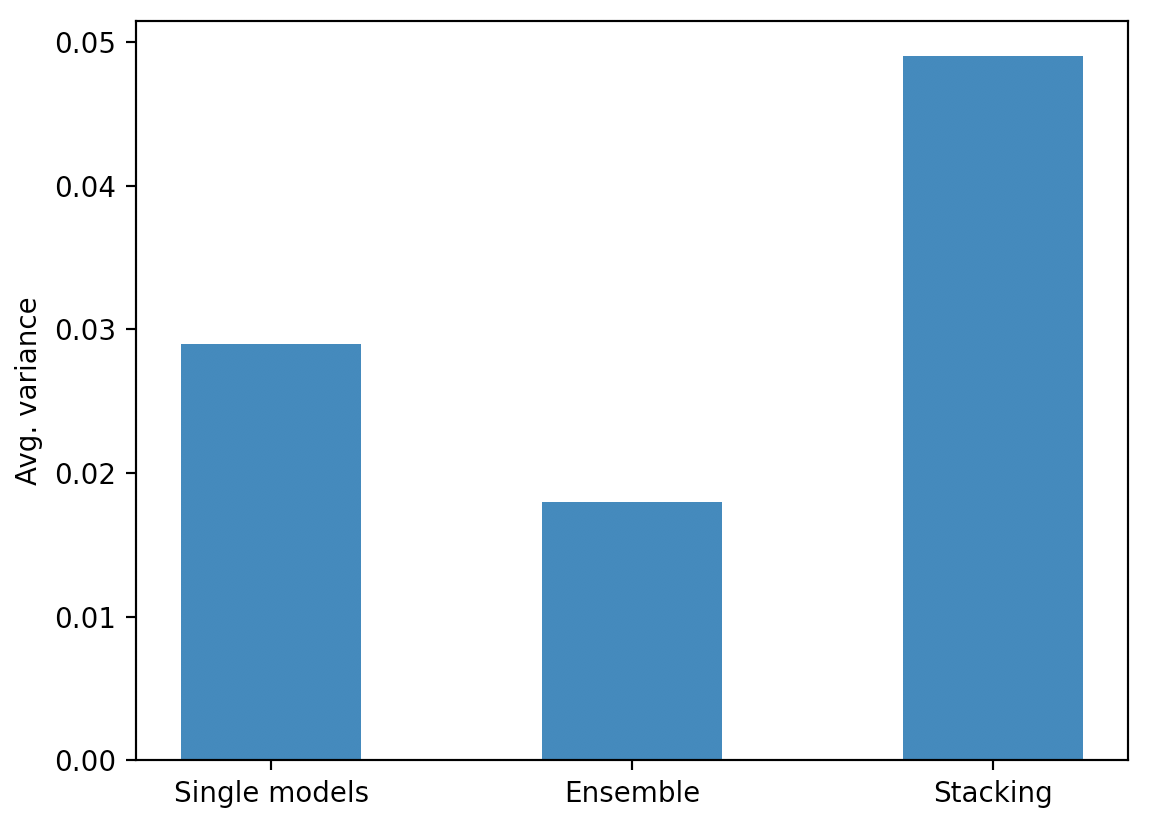}
\caption{Average variances over setups}
\label{avgvars}
\end{figure}

Given the accuracy scores, there are some additional results which are to be mentioned. Six out of eight times, the best performing model turned out to be the SVM, which is in conclusion, the most suited model for this problem. Along with the SVM, XGBoost and Ridge classifiers also appear frequently as best performing models given the different experimental setups. Important to note, that however in the majority of the cases, stacking yields the best average accuracy scores, the variance of the accuracies are the highest. The lowest average variance values belong to the ensemble setups, indicating that the confidence level rises when combining different models for predictions. Figure \ref{avgvars} shows the average variances over different experimental setups.

\begin{figure}[htb]
\centering
\scalebox{1.0}{
\includegraphics[width=1\linewidth]{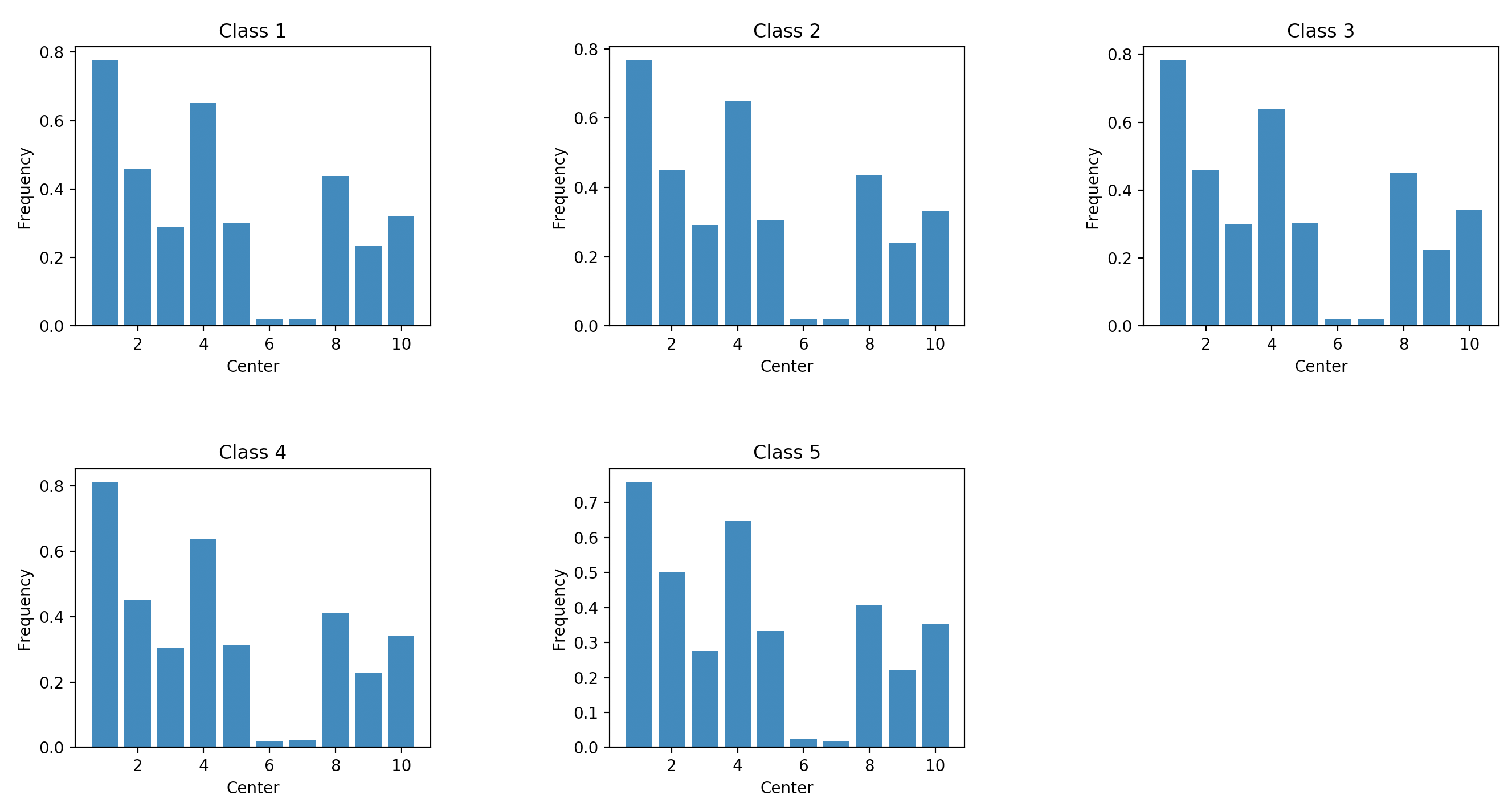}}
\caption{Average frequencies of SURF features for \textit{k = 10} cluster centers across 5 classes}
\label{h10}
\end{figure}

The least reliable performances were obtained when using the \textbf{D1} dataset, which means that the SURF features do not add to the capturing of the level of harmony. The fact that the SURF histograms show significant similarities across different classes, explains why they do not contribute to the predictions. This outcome is not surprising given the design of SURF to extract scale- and rotation-invariant interest point detectors and descriptors which is not desirable in this type of research. Figure \ref{h10} shows averaged SURF histograms for the original 5 classes with \textit{k = 10} cluster centers.

The best mean accuracy (0.80) was obtained using the \textbf{BG} setup with stacked generalization on the \textbf{D2} dataset. We can see that the ensemble models made the predictions more confident and the stacking managed to further increase mean accuracy.

\section{\uppercase{Conclusion}}
The goal of this paper is to explore whether the subjective perception of harmony can be expressed numerically. The results show that given a sufficiently large collection of randomly generated black and white images and the above described features, there exists an experimental setup by which it is possible for a machine learning model to distinguish between $Good$ and $Bad$ compositions with 80\% accuracy. However, the performance of the models decreased when all three classes were involved. That shows that when separating between classes being most distant in their level of harmony, it is possible to assign numerical values to subjectively judged compositions in order for an algorithm to confidently classify them.  Given the experimental results, we conclude that the SVM and XGBoost classifiers are the most suited for this problem.

The research described here opens several interesting future directions. Among them are, in particular, the design of more sophisticated and more expressive features, the collection and pre-processing of more data from more participants, the extension of the type and style of artistic compositions, and the exploration of different scales for rating.  Furthermore, the research introduces the need for interdisciplinary collaboration (e.g. by actively involving artists), serving as a bridge between feature design and art. 


\bibliographystyle{apalike}
{\small
\bibliography{references}}

\end{document}